# A New Knowledge Distillation Network for Incremental Few-Shot Surface Defect Detection

Chen Sun, Liang Gao, Xinyu Li and Yiping Gao

*Abstract*—Surface defect detection is one of the most essential processes for industrial quality inspection. Deep learning-based surface defect detection methods have shown great potential. However, the well-performed models usually require large training data and can only detect defects that appeared in the training stage. When facing incremental few-shot data, defect detection models inevitably suffer from catastrophic forgetting and misclassification problem. To solve these problems, this paper proposes a new knowledge distillation network, called Dual Knowledge Align Network (DKAN). The proposed DKAN method follows a pretraining-finetuning transfer learning paradigm and a knowledge distillation framework is designed for fine-tuning. Specifically, an Incremental RCNN is proposed to achieve decoupled stable feature representation of different categories. Under this framework, a Feature Knowledge Align (FKA) loss is designed between class-agnostic feature maps to deal with catastrophic forgetting problems, and a Logit Knowledge Align (LKA) loss is deployed between logit distributions to tackle misclassification problems. Experiments have been conducted on the incremental Few-shot NEU-DET dataset and results show that DKAN outperforms other methods on various few-shot scenes, up to 6.65% on the mean Average Precision metric, which proves the effectiveness of the proposed method.

## I. Introduction

Quality inspection is an important part of industrial manufacturing [1], [2]. For industrial products such as steels, printed circuit boards, and ceramics, surface defects generated during the production process not only affect appearance but also have security risks. Thus, surface defect detection is an essential part of industrial production [3]. In recent years, with the development of Internet of Things (IoT) technology, computer vision and deep learning methods have been widely used in smart manufacturing, and deep learning-based automated surface defect detection is the trend of future development [4].

Although deep learning methods can detect defect instances accurately, they need large-scale data for training and can only detect defect categories that have shown in the training stage.

Manuscript received Month xx, 2xxx; revised Month xx, xxxx; accepted Month x, xxxx. This work was supported in part by xxx and xx.

Chen Sun, Liang Gao, Xinyu Li and Yiping Gao are with Huazhong University of Science and Technology, Wuhan 430074, China

However, in real-world manufacturing, defect data are collected in the continual stream and novel categories of defects come out sequentially with only a few data. To generalize to novel data, re-training or ensemble is necessary for conventional deep learning methods, which is time-consuming and costly, with a risk of overfitting [5].

Meanwhile, incremental learning can learn from a continual data stream and extend to detect novel defect data in an end-to-end manner [6]. The novel defect data is often very few in quantity, as they occur in low frequency and take a long time to be collected [7], [8]. And detecting incremental novel defects with few-shot samples is called Incremental Few-Shot Defect Detection (IFSDD) problem. The IFSDD problem aims at learning from abundant data of base defect categories (also called old categories) and generalizing to few-shot novel defect categories. To deal with potential manufacturing issues and safety risks, all categories need to be detected, even at an early stage, thus performances on both base and novel defect categories need to be evaluated.

There are mainly two difficulties in the IFSDD problem. One is the catastrophic forgetting problem [9]. When novel defect categories are added, performance on previously base categories tends to significantly degrade over time. The other one is the misclassification problem. Industrial defect samples have complex textures yet insufficient semantic information, which causes inter-class similarity and the intra-class difference between different categories [10]. The unstable quality of few-shot data aggravates this problem and leads to misclassification.

Several methods have been proposed to tackle these difficulties in the incremental few-shot detection problem. FsdetView [11] designs a joint feature embedding module in a meta-learning framework to leverage rich feature information from base categories and generalize to novel categories, but their performances are heavily influenced by data quality and need careful adjustments according to concrete data. Some other methods [12]–[14] share a similar transfer learning paradigm, which pre-train on abundant base data and fine-tune on novel categories. Wang *et al*. [12] propose a two-stage fine-tuning approach (TFA), which keeps the parameters of the feature extraction network fixed and fine-tunes the last two fully-connected layers on a small balanced data to avoid catastrophic forgetting. Based on TFA, Sun et al. use a Few-Shot Contrastive proposal Encoding (FSCE) [13] method, which also fixes some components when fine-tuning and introduces an extra contrastive head, together with contrastive proposal encoding loss, to alleviate instance-level misclassification. Although the transfer learning paradigm with parameter fixing can maintain knowledge learned from base categories, models are limited to improve performance on novel categories. So how to achieve a balance between keeping base knowledge and exploring novel knowledge is a crucial issue for the IFSDD problem.

In this paper, a new knowledge distillation network called Dual Knowledge Align Network (DKAN) is proposed for the IFSDD problem. The proposed DKAN method also follows a transfer learning paradigm, but a teacher-student knowledge distillation framework is designed for fine-tuning, to achieve a better base knowledge retention and novel knowledge exploration. Under this framework, a novel Incremental RCNN network is proposed to decouple feature representation of base and novel categories and alleviate unstable data quality. Moreover, a feature-level distillation loss, called Feature Knowledge Align (FKA) loss is used to minimize the feature distribution gap between the teacher and student model, which alleviates catastrophic forgetting problems. An instance-level distillation loss, called Logit Knowledge Align (LKA) loss, is proposed to adjust logit distribution between the base and novel categories, which solves the misclassification problem. To evaluate the performance on the IFSDD task, the proposed method is tested on a public dataset NEU-DET under several few-shot scenes. The experimental results show that the proposed method outperforms other methods, up to 6.65% on the mean Average Precision (mAP) metric, which proves the effectiveness of the proposed method.

The remaining of this paper is organized as follows. Section II states the related work, Section III presents the detail of the proposed method. Section IV is the experimental result on the public dataset. Section V is the conclusion and future work.

## II. RELATED WORK

### A. *Defect Detection*

The purpose of the defect detection task is to classify categories and locate the position of defects. Early researchers use traditional image processing methods to extract defect features, such as histogram of oriented gradient (HOG) [15], scale-invariant feature transform [16], and speeded up robust features [17]. And classification models like support vector machine (SVM) are then used to determine categories of defects [18]. Detection results of traditional methods heavily rely on the quality of hand-craft features, which is time-consuming and can be influenced by many factors, such as illumination, defect types, and environment.

Recently, deep learning methods have been widely studied as applied in the defect defection area as they can automatically extract features from input data in an end-to-end manner[8], [19]. PGA-Net [20] designs a pyramid feature fusion and global context attention network for pixel-wise detection of the surface defect. Ni *et al.*[21] introduce a TDA training strategy and integrate CSFA-Hourglass with CASIoU-CEHM to solve the data imbalance and complex situations in rail surface defect detection. As to inter-class similarity and non-salient of flexible printed circuit boards, Luo *et al.* [22] design a decoupled two-stage detection framework, which separately conducts location and classification tasks through a multi-hierarchical aggregation module and locally non-local block. Although deep learning-based detection methods have seen great progress in different industry applications, a large bunch of data is still necessary for training. When encountering incremental few-shot problems, defect detection models fail to generalize properly on all categories.

### B. *Few-shot Defect Detection*

In recent years, few-shot learning has achieved several great breakthroughs. Many models have been proposed to solve few-shot classification tasks [23]–[26]. In contrast, few-shot object detection, as an emerging task, is less explored. Some early works follow meta-learning strategies [27], [28], which construct several meta-training tasks on base categories to learn class-agnostic feature extraction and then apply them to novel categories. FsDetView[11] designs a joint feature embedding module in a meta-learning framework to leveraging on rich feature information from base categories. Wang *et al.* [12] designed a transfer learning-based method that firstly pre-trains detection models on base categories and then freeze part parameters of the model to fine-tune on novel categories. They also claim that few-shot detection models should be evaluated on novel categories as well as base categories to examine knowledge retention. Inspired by TFA, Wu *et al.* [14] follow the transfer learning paradigm and propose a multi-scale positive sample refinement (MPSR) approach to handle scale variation problems in few-shot object detection. Furthermore, FSCE [13] demonstrates performance degradation in novel categories mainly comes from inter-class similarity. Thus, an extra contrastive head is added in the RCNN model together with a contrastive proposal encoding loss to promote inter-class compactness and inter-class variance on the instance level.

Progress in few-shot object detection on nature images promotes research on industrial images. Cheng *et al.* [29] adopt a two-phase training scheme and a metric-based detection is designed to compute distances between the representation of each class and conduct classification tasks. Wang *et al.* [7] publish a benchmark of few-shot defect detection, based on the public NEU-DET [30] dataset. For better generalization on novel categories, two domain generalization methods are developed to enhance the appearance of novel features. Although comparable results have been achieved in [7], only novel categories are evaluated for model performance, which may lead to "catastrophic forgetting" on base categories.

### C. *Incremental Learning*

Incremental learning aims at continually gathering knowledge as different tasks or data introduced without the need to retrain from scratch [6], which can also be referred as life-long learning, continual learning, or sequential learning.

Category incremental is a normal question in the real world and has been widely studied. For the incremental problem in the classification task, class word vectors, which served as auxiliary semantic information, are deployed with visual information to solve catastrophic forgetting problems in [31]. ERDIL [32] uses relation graphs to represent knowledge in different categories and design an exemplar relation loss to preserve and transfer knowledge among categories. As to incremental problems in the detection task, Fan *et al.* [33] utilize multiple RPN to capture foreground boxes for base and novel categories and design a distillation loss to measure distributions between different branches of RCNN. Chen *et al.* [34] design a new architecture for incremental object detection, which expands a new branch in the last layer of FPN, and uses knowledge distillation to maintain the old learning capability.

ONCE [35] uses CenterNet[36] as a backbone to learn the feature extractor and a per-class code generator network for the novel classes. Ganea *et al.* [37] extend MTFA to incremental MTFA (iMTFA) through learning discriminative object embeddings and then merge them into class-wise representatives.

Although catastrophic forgetting can be alleviated by incremental learning methods, novel categories are often on the few-shot scale, which brings a risk of overfitting and misclassification. Thus, it's necessary to consider a balanced knowledge transfer for incremental few-shot data.

## III. PROPOSED DKAN METHOD FOR IFSDD

### A. *Problem Definition*

In real-world manufacturing scenes, categories of defects on steel surfaces are gradually increasing as production processes. Due to low frequency and equipment limitations, it is difficult to collect abundant novel defects at the early stage. This paper aims at training a generalized detection model to detect both few-shot novel categories and abundant base categories. This task is called the IFSDD task.

Following the definition described in [27], given a defect detection dataset $D = \{(x,y) | x \in X, y \in Y\}$, where x is input image and $y = (c_i, B_i), i \in 1 \cdots N$ denotes category labels $c_i$ and bounding box coordinates $B_i$ of N defect instances in the image x. $D = D_{base} \cup D_{novel}$ where $D_{base}$ consists of a set of categories $C_{base}$ with abundant instances and annotations and $D_{novel}$ consist of another set of categories $C_{novel}$ with only K (usually less than 50) instances per category. $C_{base}$ and $C_{novel}$ have no intersection, which means $C_{base} \cap C_{novel} = \emptyset$

For incremental few-shot detection, a few-shot detector $Det(.)$ is normally pre-trained on $D_{base}$ to learn abstract representations and then generalize them to $D_{novel}$. $Det(.)$ is evaluated on a set of test datasets, including both base categories and novel categories. The goal is to optimize average precision results on both base and novel categories.

### B. *Basic Detector*

Since object detectors can roughly be described as combinations of class-agnostic feature extraction and two class-specific heads for classification and regression, the DKAN framework can be applied to most of the object detection models, such as Faster RCNN[38], YOLO[39] and RetinaNet [40]. In this work, Faster RCNN is chosen as the basic detector.

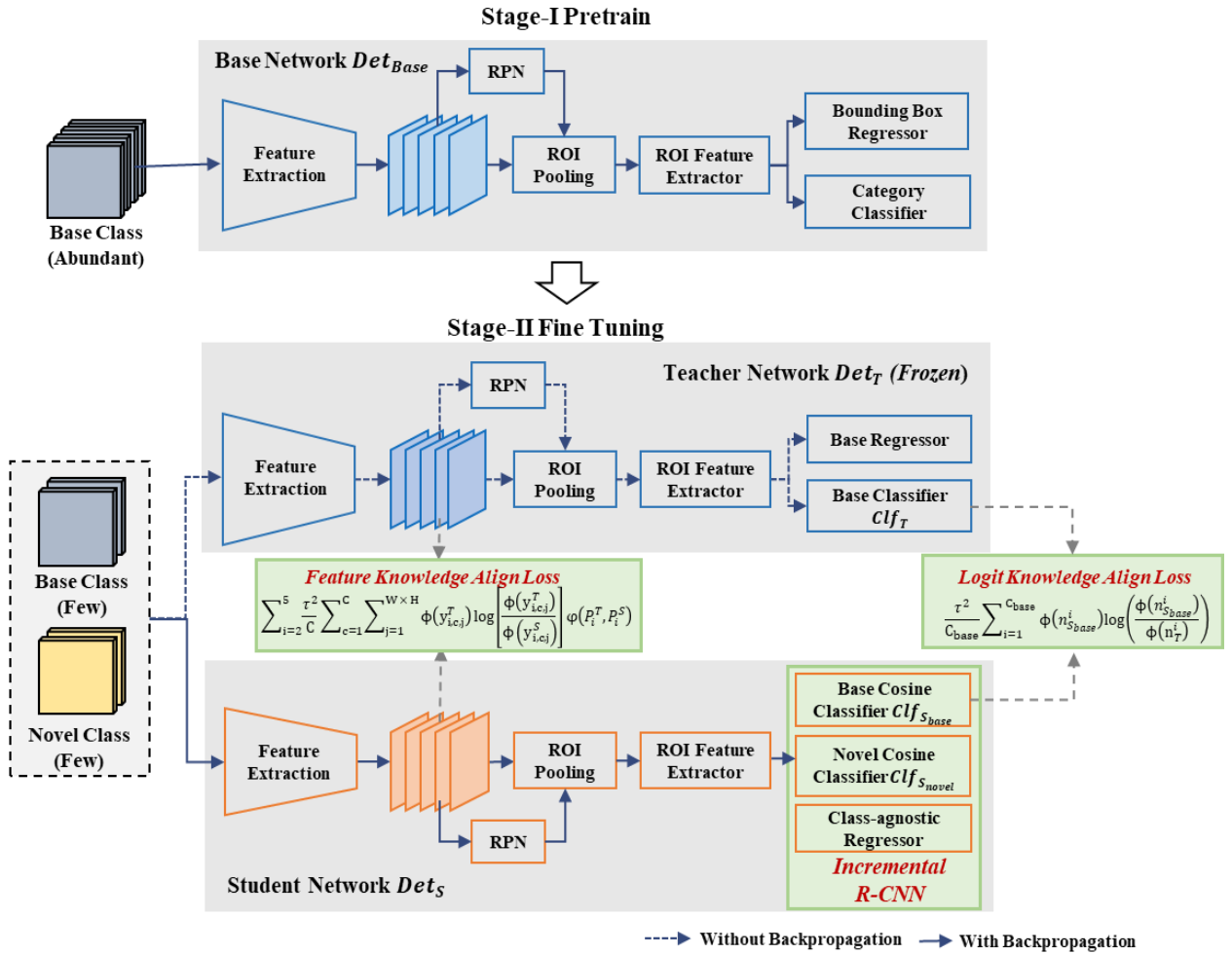

Fig. 1 Main Framework for the Dual Knowledge Align Network (DKAN) for incremental few-shot defect detection problem. The proposed DKAN method utilizes a two-stage transfer learning paradigm. In the fine-tuning stage, a Teacher-Student knowledge distillation framework is introduced. And in this framework, Incremental RCNN network is designed for alleviate unstable data quality and two knowledge align losses is proposed for knowledge updating and retention.

In Faster RCNN, the class agnostic part includes the feature extraction network, Region Proposal Network (RPN), ROI Pooling Layers, and ROI feature extractor. A feature extraction network is firstly used to extract multi-scale features from input images, which contains high-resolution feature maps for small features and low-resolution but more abstract semantic feature maps for large features. After feature extraction, RPN is applied to find foreground objects out of backgrounds and generate local candidate proposals. These candidate proposals, together with feature maps gained from the feature extraction network are then sent into ROI Pooling and ROI Feature Extractor to extract fixed-size feature vectors.

As to class-specific heads, normally two fully-connected layers are deployed as the Instance Classifier and Bounding Box Regressor. They take feature vectors from ROI Feature Extractor as input and output categories of instances and coordinates of bounding boxes in the images.

In this work, class agnostic parts are generally denoted as $F(\cdot)$, the classification head is denoted as $Clf(\cdot)$ and the regression head is denoted as $Reg(\cdot)$.

### C. Main Framework of the Proposed Method

The proposed Dual Knowledge Align Network (DKAN) utilizes a two-stage training paradigm. As shown in Fig. 1, in the stage-I, a detector $Det_{base}$ is pretrained on base categories with abundant instances and annotations. By pretraining, $Det_{base}$ learns abstract representation from base categories.

In stage II, to generalize to novel categories, a teacher-student knowledge distillation framework is introduced, including a teacher network $Det_T$ and a student network $Det_S$. The teacher network $Det_T$ copies the structure and parameters from the pre-train network $Det_{base}$ and will not update parameters in the following fine-tuning stage, which is noted in Fig. 1 as "Frozen". Therefore, the teacher network $Det_T$ preserves base-class knowledge learned from the pre-train stage. Meanwhile, the student network has a different structure. To detect both base and novel categories in an end-to-end manner, an Incremental RCNN structure is designed in the student network $Det_S$. This Incremental RCNN structure decouples feature representation for base and novel categories. And two Cosine-Similarity classifiers are deployed in the incremental RCNN to alleviate the influence of unstable few-shot data.

To balance the knowledge updating and retention process, two knowledge align losses, Feature Knowledge Align (FKA) and Logit Knowledge Align (LKA), are designed between the teacher network $Det_T$ and the student network $Det_S$. The FKA loss transfers knowledge of base categories from $Det_T$ to $Det_S$ by aligning the distribution of class-agnostic feature maps between two models, which is a "softer" way compared to the direct parameter freeze process. The LKA loss adjusts the distribution of classification logits of the student network $Det_S$ according to the logit output of the teacher network $Det_T$, which deals with the misclassification problem in few-shot data.

The overall loss in the fine-tuning stage is shown below.

$$Loss_{all} = Loss_{RPN} + Loss_{RCNN} + \lambda_1 Loss_{FKA} + \lambda_2 Loss_{LKA} \quad (1)$$

where $Loss_{RPN}$ and $Loss_{RCNN}$ are the original losses in Faster RCNN and their settings follow the one in [38]. $\lambda_1$ and $\lambda_2$ are the loss weights for $Loss_{CFA}$ and $Loss_{RLA}$.

### D. Proposed Incremental R-CNN Network

With novel categories added in fine-tuning stage, dimensions of the classification and regression head need to expand. Thus, an Incremental RCNN network is designed in $Det_S$.

For regression, the proposed Incremental RCNN utilize a class-agnostic regression head to conduct localization task. For classification, besides classification head $Clf_{S_{base}}$ for base categories inherited from $Det_{Base}$, an extra head $Clf_{S_{novel}}$ for novel categories is introduced, thus decoupling the feature representation between base and novel categories.

Cosine-Similarity classifiers, instead of normally adopted fully-connected layers, are used in $Clf_{S_{base}}$ and $Clf_{S_{novel}}$. Compared to fully-connected layer, instance-wised feature normalization is applied before classification in a cosine similarity-based classifier, which is effective for few-shot data because few-shot data collected in real scenes have unstable data quality and can aggravate intra-class difference and inter-class similarity between defect categories and brings misclassification problem. [12].

The Cosine Similarity-based classifier is defined as follows:

$$c_{i,j} = \alpha \cdot \frac{f_i \cdot w_j}{\|f_i\| \|w_j\|} \quad (2)$$

where $f_i$ is the input feature of category i and $w_j$ is the weight parameter for category j, $\alpha$ is the scaling factor. $c_{i,j}$ measures the similarity between feature $f_i$ and category vector $w_j$.

### E. Feature Knowledge Align Loss

In transfer-learning-based few-shot detection methods, like TFA [12] and FSCE [13], base knowledge retention is realized by freezing the whole or parts of parameters in the class-agnostic part. However, the "freeze" operation may fail to update feature representations from novel categories. In this part, a Feature Knowledge Align (FKA) loss is introduced to provide a "softer" way of knowledge transfer.

As shown in Fig. 2, the proposed FKA loss is utilized between class-agnostic feature maps of $Det_S$ and $Det_T$. These feature maps are the outputs of the FPN network, noted as $\{P_2, P_3, P_4, P_5\}$ specifically in Faster RCNN.

As mentioned before, the teacher network $Det_T$ preserves knowledge from base categories and the student network $Det_S$ explores knowledge in novel categories. By aligning the distribution of feature maps, the student network's class-agnostic part $F_S$ can learn new representations of novel categories without forgetting knowledge from base categories.

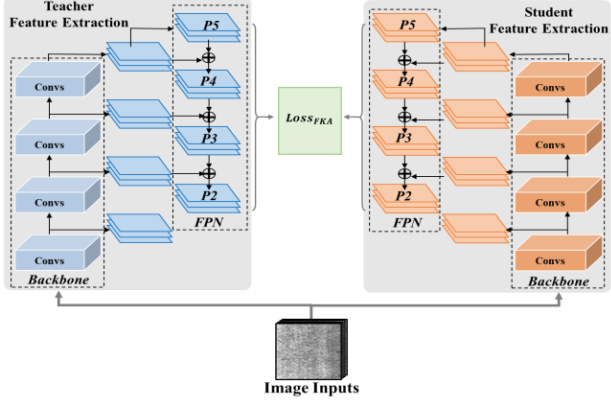

Fig. 2 The implementation framework of Feature Knowledge Align loss

As activation of different channels tends to contain different levels of semantic information [41], a channel-wise loss is used to realize feature map distribution alignment. Firstly, activations of feature maps are converted into a probability distribution in convenience to measure discrepancy, the equation is below.

$$\phi(y_c) = \frac{\exp\left(\frac{y_{c,i}}{\tau}\right)}{\sum_{i=1}^{W \times H} \exp\left(\frac{y_{c,i}}{\tau}\right)} \quad (3)$$

where $y_{c,i}$ is the channel value. $c = 1, 2, \ldots, C$ is the channel index and i is the spatial location of channels. $\tau$ is the hyperparameter of temperature. A larger $\tau$ brings a wider spatial location.

After the conversion of the feature map, KL divergence is used to measure channel distribution between feature maps in $Det_S$ and $Det_T$, the equation is below.

$$\varphi(P_i^T, P_i^S) = \frac{\tau^2}{C} \sum_{c=1}^{C} \sum_{j=1}^{W \times H} \phi(y_{i,c,j}^T) \log\left[\frac{\phi(y_{i,c,j}^T)}{\phi(y_{i,c,j}^S)}\right] \quad (4)$$

where $P_i^T$ and $P_i^S$ are the feature maps in $Det_T$ and $Det_S$, $y_{i,c,j}^T$ and $y_{i,c,j}^S$ are the channel values in $P_i^T$ and $P_i^S$. $\varphi$ is the KL divergence between feature maps. It can be seen that as $\phi(y_{i,c,j}^T)$ gets larger, $\phi(y_{i,c,j}^S)$ needs to be as large as $\phi(y_{i,c,j}^T)$ to minimize $\varphi$, which means similar activation of feature maps would be passed from the teacher model to the student model.

The final FKA loss is calculated by summing up KL divergence values on 4 different feature maps, $\{P_2, P_3, P_4, P_5\}$, and the equation is shown below

$$Loss_{FKA} = \sum_{i=2}^{5} \varphi(P_i^T, P_i^S) \quad (5)$$

F. *Logit Knowledge Align Loss*

As the teacher model $Det_T$ directly copies the structure and parameters of the pre-trained model $Det_{base}$, classification head $Clf_T$ in the teacher model tend to have a similar response as $Det_{base}$, which means it would have high activation for base categories and low activation for novel categories. For the student model, decoupled cosine classifier $Clf_{S_{base}}$ and $Clf_{S_{novel}}$ conduct classification tasks for base and novel categories separately. It is believed that base classification head $Clf_{S_{base}}$ in the student model ought to have similar output distribution as $Clf_T$ while novel classification head $Clf_{S_{novel}}$ is supposed to have an output distribution different from $Clf_T$.

Based on this, a Logit Knowledge Align (LKA) is proposed to adjust the output logit distribution of student model $Det_S$. The LKA loss directly measures the distribution distances between logit outputs of $Clf_{S_{base}}$ and $Clf_T$, as shown in Fig. 3.

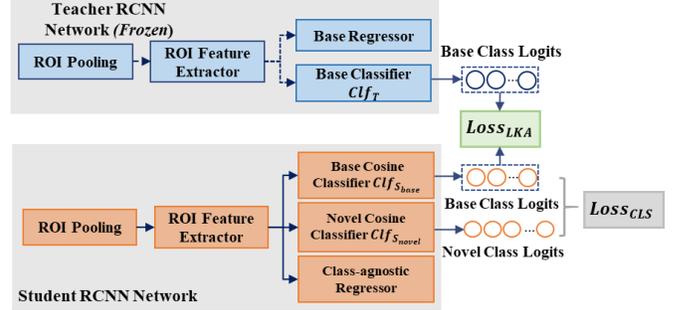

Fig. 3 The implementation framework of Logit Knowledge Align loss

To align classification outputs between $Clf_{S_{base}}$ and $Clf_T$, base logit outputs are firstly converted into a probability using the softmax function. Equations are shown below.

$$\phi(n_T^i) = \frac{\exp\left(\frac{n_T^i}{\tau}\right)}{\sum_{i=1}^{C_{base}} \exp\left(\frac{n_T^i}{\tau}\right)} \quad (5)$$

$$\phi(n_{S_{base}}^i) = \frac{\exp\left(\frac{n_{S_{base}}^i}{\tau}\right)}{\sum_{i=1}^{C_{base}} \exp\left(\frac{n_{S_{base}}^i}{\tau}\right)} \quad (6)$$

where $i = 1, 2, \ldots, C_{base}$ is the category index of base categories, $n_T^i$ and $n_{S_{base}}^i$ are logit outputs on base category i from $Clf_T$ and $Clf_{S_{base}}$. $\tau$ is the hyperparameter of temperature.

Then KL divergence is deployed to measure the distribution distance. The final equation for the LKA loss is as follows.

$$L_{LKA} = \frac{\tau^2}{C_{base}} \sum_{i=1}^{C_{base}} \phi(n_{S_{base}}^i) \log\left(\frac{\phi(n_{S_{base}}^i)}{\phi(n_T^i)}\right) \quad (7)$$

IV. EXPERIMENTS

A. *Dataset Description*

To evaluate performance on the incremental few-shot defect detection task, NEU-DET [30], a public defect detection dataset, is utilized and transformed into Incremental Few-shot NEU-DET dataset(I-FSND).

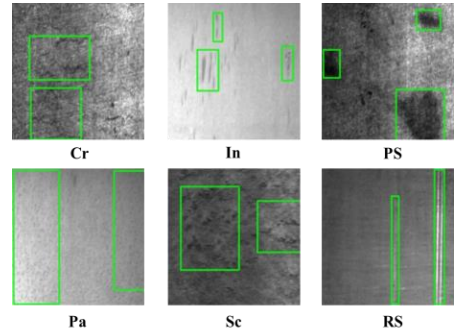

Fig. 4. Examples of steel surface defects in NEU-DET. Defect areas are marked in green boxes.

The original NEU-DET datasets collect 6 types of defects on hot-rolled steel strips, including Crazing (Cr), Inclusion (In), Pitted Surface (PS), Patches (Pa), Scratches (Sc), and Roll-in Scales (RS). There are 1800 gray-scale images in total. And each category has 300 samples. The original resolution of images is $200 \times 200$. Examples of the NEU-DET dataset are shown in Fig. 4, and ground truth defect areas are marked in green boxes.

In the I-FSND dataset, six defect categories in NEU-DET are divided into base categories $C_{base}$ and novel categories $C_{novel}$. The data split process is shown in Fig. 5. For model training and evaluation, testing data $D^{test} = (D^{test}_{base}, D^{test}_{novel})$ is randomly chosen from original data and each category contains 60 images. Meanwhile, for training data $D^{train} = (D^{train}_{base}, D^{train}_{novel})$, $D^{train}_{base}$ contains 240 images for each category in $C_{base}$ while $D^{train}_{novel}$ only has K images per category (K is less than 50 normally). Each novel image only has one labeled instance.

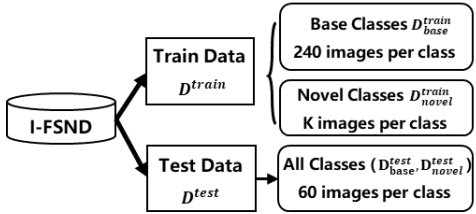

Fig. 5 Data split process in I-FSND dataset.

To evaluate the performance of models under different few-shot scenes, as shown in TABLE I, 3 base-novel splits are given, and K-shot number in $D^{train}_{novel}$ varies in 5,10,30.

TABLE I BASE-NOVEL CATEGORY SPLIT RESULTS

|  | **Base Class** | **Novel Class** | **Novel Shots** |
|---|---|---|---|
| ***SPLIT 1*** | Cr, In, PS | Pa, Sc, RS |  |
| ***SPLIT 2*** | In, RS, Sc | Cr, Pa, PS | 5/10/30 |
| ***SPLIT 3*** | PS, Pa, Sc | Cr, In, RS |  |

### B. Experimental Details and Evaluation Metrics

The code of the DKAN method is implemented with PyTorch1.10 and MMDetection toolbox [42]. All the experiments are performed on NVIDIA RTX 3090 GPU (with 24G memory) under CUDA11.1 on Ubuntu 16.04 system. The size of the input image is $800 \times 800$. The backbone network used in experiments is pre-trained on the ImageNet classification task. And The default setting of the backbone network is Faster RCNN 101 with FPN. For the training process, the batch size is set as 4, the number of total training iterations is set as 2000, and the learning rate is set as 0.02 with a decay of 0.0001. Others follow default settings in the MMDetection toolbox.

The effect of detection is evaluated by Average Precision (*AP*), a comprehensive metric used in object detection. AP calculates areas under the P-R curve, where both Precision and Recall are evaluated. And in this work, AP with a threshold value of 50%, also noted as AP50, is used to evaluate performance on base categories, novel categories, and all categories noted as ***AP**B*, ***AP**N*, and ***AP**All*.

$$\text{Precision} = \frac{TP}{TP+FP} \quad (8)$$

$$\text{Recall} = \frac{TP}{TP+FN} \quad (9)$$

$$\text{AP} = \int_0^1 P(R)dR \quad (10)$$

### C. Influence of the Combination of Loss Functions

This section explores the influence of two loss functions $\text{Loss}_{FKA}$ and $\text{Loss}_{LKA}$ in knowledge distillation framework. Especially, the influence of two loss functions is studied under SPLIT-1 5 shot in TABLE I.

Firstly, the optimal loss weight of $\text{Loss}_{FKA}$ and $\text{Loss}_{LKA}$ are searched. Concretely, a two-step search process is deployed. $\lambda_1$ is searched independently in the scope of $\{0.001, 0.01, 0.1, 1, 10\}$ and as shown in Fig. 6(a), the best results are achieved when $\lambda_1 = 1$. Then $\lambda_2$ is also searched in the scope of $\{0.001, 0.01, 0.1, 1, 10\}$ with $\lambda_1 = 1$. As shown in Fig. 6(b), the best results are achieved when $\lambda_2 = 0.01$.

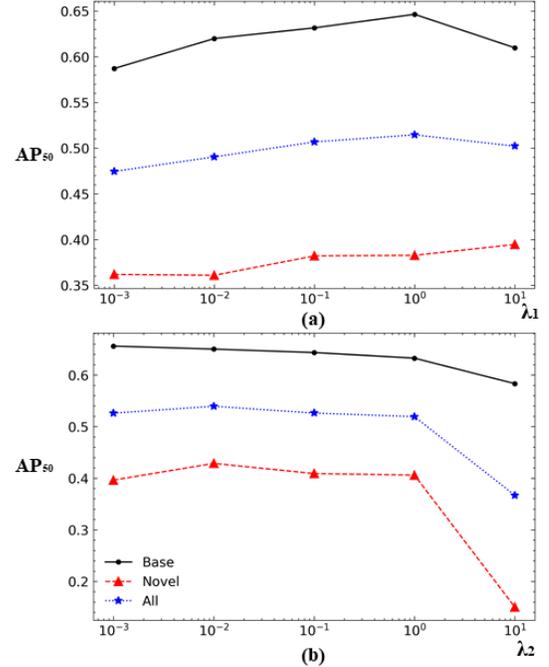

Fig. 6 The two-step search for optimal settings of loss weights. (a) The performance of loss weight $\lambda_1$ with $\lambda_2 = 0$. (b) The performance of loss weight $\lambda_2$ with best $\lambda_1$ setting obtained in (a)

$\text{Loss}_{FKA}$ and $\text{Loss}_{LKA}$ play different roles in the knowledge distillation framework. Ablation experiments are conducted and results are shown in TABLE II. Without the constraint of $\text{Loss}_{FKA}$ and $\text{Loss}_{LKA}$, DKAN achieves 0.4745 on ***AP**All* metric. When only adding $\text{Loss}_{FKA}$, ***AP**B* is improved by 5.93% while ***AP**N* is improved by 2.09%. Meanwhile, when only adding $\text{Loss}_{LKA}$, ***AP**B* is improved by 5.24% and ***AP**N* is improved by 5.71%. If applying both, a comprehensive improvement is seen on both ***AP**B* and ***AP**N*, achieving a new highest result of 42.86% on ***AP**N*.

TABLE II THE INFLUENCE OF LOSS FUNCTIONS

| $L_{FKA}$ | $L_{LKA}$ | $AP_B$ | $AP_N$ | $AP_{All}$ |
|---|---|---|---|---|
|  |  | 0.5871 | 0.3618 | 0.4745 |
| ✓ |  | 0.6464 | 0.3827 | 0.5145 |
|  | ✓ | 0.6395 | 0.4189 | 0.5292 |
| ✓ | ✓ | 0.6501 | 0.4286 | 0.5394 |

This ablation study shows that $\text{Loss}_{FKA}$ transfers the global feature knowledge from $\text{Det}_T$ to $\text{Det}_S$, and improves results on base categories. As a complement, Loss$_{LKA}$ improves detection results on novel categories by adjusting logit distribution and also brings an improvement on base categories, although not as good as Loss$_{FKA}$. In the end, a weighting combination of both losses brings a better result than a single addition of either one.

A similar conclusion can be drawn from where detailed classification ratio results of 6 categories and backgrounds are shown using a confusion matrix. The diagonal of the matrix is the true positive ratio in each class. Compared to the model in Fig. 7(a) with no distillation loss, the DKAN method with $L_{FKA}$ shows better classification accuracy on base categories in Fig. 7 (c) and the DKAN method with $L_{LKA}$ greatly improves true positive ratio on novel categories as shown in Fig. 7(b), especially in the category "Sc" which has been improved by 23%. And a combination on Fig. 7(d) brings a comprehensive improvement on all categories.

Fig. 7 Confusion Matrix comparison between different combinations of loss functions

### D. Influence of the Loss Hyperparameters τ

This part studies the influence of hyperparameter τ in both FKA and LKA loss. The hyperparameter τ plays an important role in controlling the width of knowledge to be transferred. And five values, including 3, 4, 5, 6, and 7, are chosen for study. Experiments are conducted under SPLIT-1 5-shot scene.

Results are shown in TABLE IV. As the value of τ increases, both $AP_B$ and $AP_N$ first increase and then decrease, the best result of $AP_B$ is achieved when τ = 4, and the best result of $AP_N$ is achieved when τ = 5. The value of τ = 5 achieves the best results on $AP_{All}$.

TABLE IV THE INFLUENCE OF HYPERPARAMETER τ

|   | $AP_B$ | $AP_N$ | $AP_{All}$ |
|---|---|---|---|
| τ = 3 | 0.6114 | 0.3811 | 0.4963 |
| τ = 4 | 0.6519 | 0.3923 | 0.5221 |
| τ = 5 | 0.6501 | 0.4286 | 0.5394 |
| τ = 6 | 0.6499 | 0.3942 | 0.5221 |
| τ = 7 | 0.6194 | 0.3880 | 0.5037 |

### E. Influence of proposed Incremental RCNN

This section explores the influence of the proposed Incremental RCNN, including the cosine similarity classifier and its hyperparameter α. Comparisons are conducted with normal fully-connected layers and different values for hyperparameter α are also studied, including 5, 10, 20, and 50. All the results are under the SPLIT-1 5 shot.

TABLE V INFLUENCE OF COSINE SIMILARITY CLASSIFIER

|   |   | $AP_B$ | $AP_N$ | $AP_{All}$ |
|---|---|---|---|---|
| DKAN (w / fc)* | | 0.6211 | 0.3148 | 0.4679 |
| DKAN (w / cos)** | α = 5 | 0.6421 | 0.4443 | 0.5432 |
|  | α = 10 | 0.6571 | 0.4211 | 0.5391 |
|  | α = 20 | 0.6501 | 0.4286 | 0.5394 |
|  | α = 50 | 0.6174 | 0.4043 | 0.5109 |

*(w /fc) means DKAN with fully connected layers
**(w /cos) means DKAN with cosine similarity classifier

Results are shown in TABLE V. It can be seen that DKAN (w./ cos) generally have better AP results than DKAN (w./ fc). The cosine similarity classifier brings a huge performance improvement on $AP_N$, up to 13.95% when α = 5. This comparison shows the cosine similarity classifier is effective.

TABLE III COMPARISON WITH STATE-OF-THE-ART FEW-SHOT DETECTION METHODS

|  | Methods | SPLIT 1 | | | SPLIT 2 | | | SPLIT 3 | | |
|---|---|---|---|---|---|---|---|---|---|---|
|  |  | $AP_B$ | $AP_N$ | $AP_{All}$ | $AP_B$ | $AP_N$ | $AP_{All}$ | $AP_B$ | $AP_N$ | $AP_{All}$ |
| 5 shot | FsDetView | 0.6256 | 0.1282 | 0.3769 | 0.6219 | 0.1174 | 0.3696 | 0.6891 | 0.0376 | 0.3634 |
|  | TFA | 0.5947 | 0.1800 | 0.3873 | 0.6156 | 0.0070 | 0.3113 | **0.8205** | 0.1342 | 0.4774 |
|  | FSCE | 0.5860 | 0.3596 | 0.4729 | 0.6404 | 0.2279 | 0.4341 | 0.7760 | 0.1652 | 0.4706 |
|  | MPSR | 0.4228 | 0.2830 | 0.3529 | 0.5067 | **0.3813** | 0.4441 | 0.6400 | 0.1621 | 0.4011 |
|  | **Ours (DKAN)** | **0.6501** | **0.4286** | **0.5394** | **0.7313** | 0.2579 | **0.4946** | 0.8042 | **0.1963** | **0.5003** |
| 10 shot | FsDetView | 0.6339 | 0.2968 | 0.4654 | 0.7064 | 0.1677 | 0.4371 | 0.7268 | 0.1160 | 0.4214 |
|  | TFA | 0.6209 | 0.1379 | 0.3794 | **0.7175** | 0.0150 | 0.3663 | **0.8213** | 0.1571 | 0.4892 |
|  | FSCE | 0.6281 | 0.3916 | 0.5098 | 0.7073 | 0.2568 | 0.4820 | 0.7936 | 0.1915 | 0.4926 |
|  | MPSR | 0.4051 | 0.3536 | 0.3793 | 0.5443 | **0.3727** | 0.4587 | 0.6474 | 0.1903 | 0.4189 |
|  | **Ours (DKAN)** | **0.6355** | **0.4878** | **0.5617** | 0.7158 | 0.3224 | **0.5191** | 0.8054 | **0.1986** | **0.5020** |
| 30 shot | FsDetView | 0.6361 | 0.4492 | 0.5427 | 0.6979 | 0.2518 | 0.4749 | 0.7904 | 0.2510 | 0.5207 |
|  | TFA | 0.6466 | 0.1474 | 0.3970 | **0.7389** | 0.0243 | 0.3816 | 0.8296 | 0.2174 | 0.5235 |
|  | FSCE | 0.6232 | 0.3852 | 0.5042 | 0.6814 | 0.2248 | 0.4531 | 0.8034 | 0.1615 | 0.4825 |
|  | MPSR | 0.4697 | 0.4794 | 0.4745 | 0.6013 | **0.4033** | 0.5023 | 0.6755 | 0.2863 | 0.4809 |
|  | **Ours (DKAN)** | **0.6474** | **0.5025** | **0.5750** | 0.7364 | 0.3623 | **0.5494** | **0.8344** | **0.3231** | **0.5788** |

The influences of parameter α are also shown in TABLE V. When α goes bigger, the overall performance generally decreases. The best of $AP_N$ is achieved when α = 5 and the best of $AP_B$ is achieved when α = 10. The default value α = 20 makes a tradeoff between $AP_B$ and $AP_N$.

Fig. 8 utilizes a confusion matrix to show detailed classification results between 6 categories and backgrounds in DKAN (w./ fc) and DKAN (w./ cos). It can be seen from the comparison of (a) and (b), the introduction of a cosine similarity classifier improves the true positive ratio of all defect categories, 7%, 5%, 2%, 1%, 11%, and 3% for Cr, In, PS, Pa, Sc, and RS separately. These improvements are mainly due to the normalization process in the cosine similarity classifier, which generates stable feature representation and alleviates misclassification problems.

![Confusion matrices]

Fig. 8 Confusion matrix comparison between different DKAN methods on SPLIT-1 5 shot. (a) DKAN (w./ fc), (b) DKAN (w./ cos) with =20. BG is short for background.

### F. Comparison with State-of-the-art Methods

This section presents the comparison results between the proposed methods and four state-of-the-art methods in the few-shot detection task, including the meta-learning method FsDetView [11], fine-tuning methods TFA [12], FSCE [13] and MPSR [14]. All four models keep the same training settings as DKAN.

The comparisons are conducted under 3 different base-novel category splits shown in TABLE I. Each base-novel split has 3 shots for novel categories, including 5,10, and 30 shots. For a fair comparison, novel shots are sampled with 10 different random seeds. And the average values are calculated as final results.

Results are shown in TABLE III and the best mAPs are marked in bold. It can be seen that the proposed DKAN method has the best $AP_{All}$ results on all 9 shot scenes under three different base-novel splits. In detail, DKAN outperforms the second-best result by 6.65%, 5.19% and 3.23% on SPLIT-1, 5.05%, 3.761% and 4.71% on SPLIT-2, 2.29% 0.94%, 5.53% on SPLIT-3. These results indicate that the proposed method is better for the incremental few-shot defect detection task.

Moreover, as to detailed detection performance on base and novel datasets, for all 9 few-shot scenes, the proposed DKAN method achieves 5 best results on the $AP_B$ metric and 6 best results on the $AP_N$ metric. For cases where DKAN fails to achieve the best results on base or novel datasets, its performance is still competitive and comprehensively good. Taking SPLIT-2 10 shot as an example, TFA has the best $AP_B$ result and MPSR has the best $AP_N$ result. DKAN has the second-best $AP_B$ and $AP_N$ result. In a comprehensive view, DKAN outperforms TFA on $AP_N$ by 30.47% and outperforms MPSR on $AP_B$ by 17.15%. That means although TFA and MPSR achieve the best score on one metric, they sacrifice the performance on the other. Meanwhile, the proposed DKAN method manages to keep a balance between bass categories and novel categories, which proves that the DKAN method alleviates the catastrophic forgetting problem and misclassification problem.

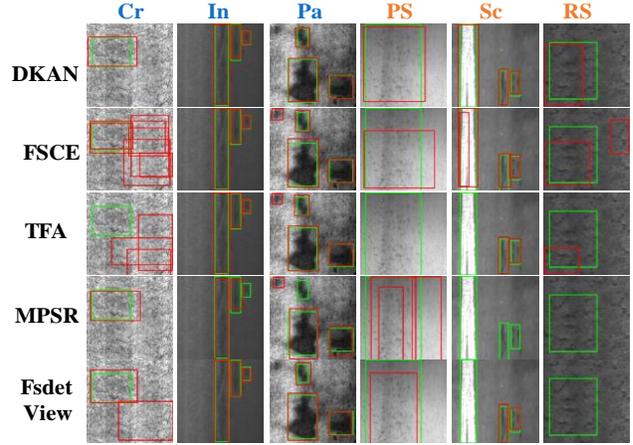
(a) SPLIT 1

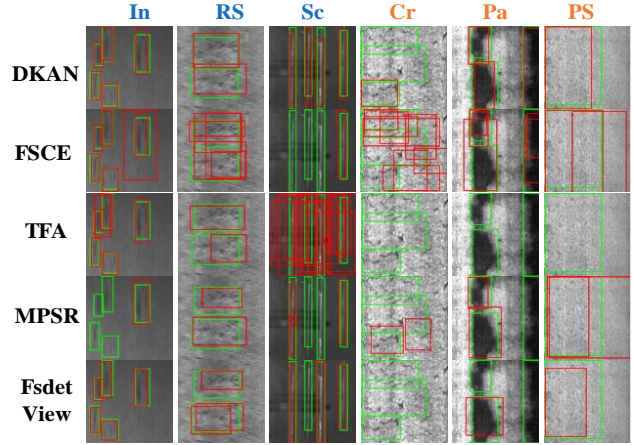
(b) SPLIT 2

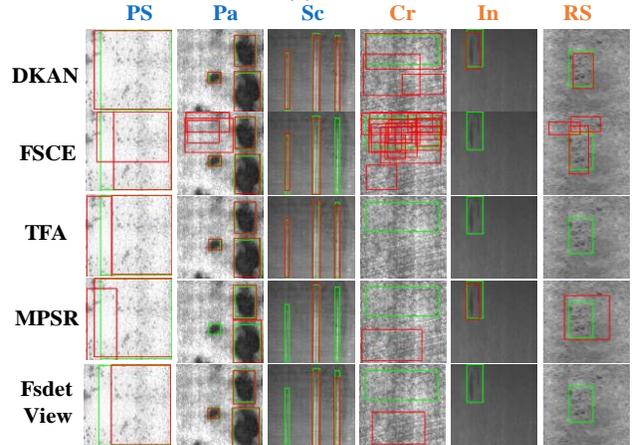
(c) SPLIT 3

Fig. 9 Visualization comparisons between DKAN and other 4 few-shot detection methods under three data splits. Among the defect category titles, Blue is for base classes and Orange is for novel classes. In the defect images, Red boxes are prediction results with confidence above 0.3 and Green boxes are ground truth boxes.

Fig. 9 shows visualization comparisons between DKAN and the other 4 few-shot detection methods. All 3 data splits are taken into consideration. For base categories, it can be seen that the proposed DKAN method can accurately classify and locate bounding boxes under all 3 data splits. Meanwhile, TFA and FSCE, which directly freeze model parameters, also perform well on base categories but still have several false positive cases and mistake the background for a defect region, such as category Cr in SPLIT 1, In in SPLIT 2. MPSR and FsdetView also show low recall cases in base categories and miss some ground truth defect boxes, such as category Sc in SPLIT 2 and SPLIT 3. These missing cases are mainly caused by catastrophic forgetting. For novel categories, the proposed DKAN method can generalize well and detect most of the defect regions. However, TFA, FSCE MPSR, and FsdetView all show some bad generalization cases. TFA even has no high confidence detection results on novel categories in SPLIT 2 and SPLIT 3. These visual comparisons suggest that the teacher-student framework in the DKAN method is effective. which alleviates the catastrophic forgetting problem and successfully explores feature representations of novel categories.

## V. Conclusion

In this paper, a new knowledge distillation network called Dual Knowledge Align Network (DKAN) is proposed for the incremental few-shot defect detection task. The proposed method uses a pretraining-finetuning transfer learning paradigm and a knowledge distillation framework is designed in the fine-tuning stage. An Incremental RCNN network is designed to achieve decoupled stable feature representation for different categories, and realize end-to-end class incremental detection. Under this framework, a feature knowledge aligns loss is proposed to align global feature distribution and alleviate the catastrophic forgetting problem. A logit knowledge aligns loss is proposed to adjust logit distribution and solve the misclassification problem. Experiments are conducted on the public dataset NEU-DET under several few-shot scenes. And the proposed DKAN achieves better performance compared with state-of-the-art few-shot detection methods, up to 6.65% on mAP metric, which proves that the proposed method is effective for the incremental few-shot defect detection task.

The limitations of the proposed DKAN methods include the following aspects. First, the proposed method is in a supervised manner, but there are lots of unlabeled defect images in the real world that have not been explored. Second, although Faster-RCNN is the baseline model in detection tasks, transformer-based algorithms have shown great potential for visual and multimodal tasks in recent years. Thus, the future work of this paper will focus on two directions. One is to develop semi-supervised or self-supervised few-shot detection algorithms. Another one is to apply transformer models to few-shot detection tasks